\documentclass[journal]{IEEEtran}

\usepackage{cite}
\usepackage{amsmath,amsfonts,amssymb}
\usepackage{algorithmic}
\usepackage{algorithm}
\usepackage{graphicx}
\usepackage{textcomp}
\usepackage{xcolor}
\usepackage{multirow}
\usepackage{url}

\usepackage[caption=false,font=normalsize,labelfont=sf,textfont=sf]{subfig}



\usepackage{pifont}

\begin{document}
\title{FedPhD: Federated Pruning with Hierarchical Learning of Diffusion Models}

\author{Qianyu Long,
        Qiyuan Wang,
        Christos Anagnostopoulos,
        and Daning Bi%
\thanks{Q. Long, Q. Wang, and C. Anagnostopoulos are with the School of Computing Science, University of Glasgow, Glasgow, UK (e-mail: q.long.1@research.gla.ac.uk; qiyuan.wang@glasgow.ac.uk; christos.anagnostopoulos@glasgow.ac.uk).}%
\thanks{D. Bi is with the College of Finance and Statistics, Hunan University, Changsha, China (e-mail: daningbi@hnu.edu.cn).}%
\thanks{This paper is currently under review at IEEE Transactions on Cybernetics.}}

\maketitle

\begin{abstract}
Federated Learning (FL), as a distributed learning paradigm, trains models over distributed clients' data. 
FL is particularly beneficial for distributed training of Diffusion Models (DMs), which are high-quality image generators that require diverse data. However, challenges such as high communication costs and data heterogeneity persist 
in training DMs similar to training Transformers and Convolutional Neural Networks. Limited research has addressed these issues in FL environments. To address this gap and challenges, we introduce a novel approach, FedPhD, designed to efficiently train DMs in FL environments. FedPhD leverages Hierarchical FL with homogeneity-aware model aggregation and selection policy to tackle data heterogeneity while 
reducing communication costs. 
The distributed structured pruning of FedPhD enhances computational efficiency 
and reduces model storage requirements in clients. 
Our experiments across multiple datasets demonstrate that FedPhD achieves high model 
performance regarding Fréchet Inception Distance (FID) scores while reducing communication costs by up to $88\%$.
FedPhD outperforms baseline methods achieving at least a $34\%$ improvement in FID, while utilizing only $56\%$ of the total computation and communication resources.
\end{abstract}

\begin{IEEEkeywords}
Federated learning, diffusion models, model pruning, hierarchical learning, privacy-preserving machine learning.
\end{IEEEkeywords}

\section{Introduction}
Diffusion Models (DMs) \cite{Ho2020,Song2021,Song2021a,Rombach2022} 
evidenced advancements in generating high-quality images compared to other generative models such as Generative Adversarial Networks (GANs) and Variational Auto-encoders (VAEs). 
Due to the flexibility and power of DMs, 
DMs have been used in many areas including computer vision, natural language generation, temporal data modeling, and multi-model learning \cite{Yang2023}. The Federated Learning (FL) paradigm originated in \cite{McMahan2017} has gained substantial attention in distributed learning by connecting \textit{data islands} with collaborative training over distributed clients, e.g., edge devices, roadside units, edge micro-servers, while preserving privacy. 
For instance, FL-based Google's Gboard learns new words and phrases without exposing clients' privacy.

With the development of DMs and FL, their integration provided a two-fold benefit \cite{Zhuang2023}. FL mitigates data scarcity during DMs training by engaging distributed clients at the network edge or organizations \cite{Zhuang2023}. 
On the other hand, DMs are powerful for high-resolution image generation, which can be deployed as data augmentation techniques to cope with data heterogeneity and high communication costs for downstream tasks in FL \cite{Huang2024a,Li2024,Ma2024,Luca2022,Zhang2022,Yang2024}. 

\begin{figure}[!ht]
    \centering
    \includegraphics[width=0.8\linewidth]{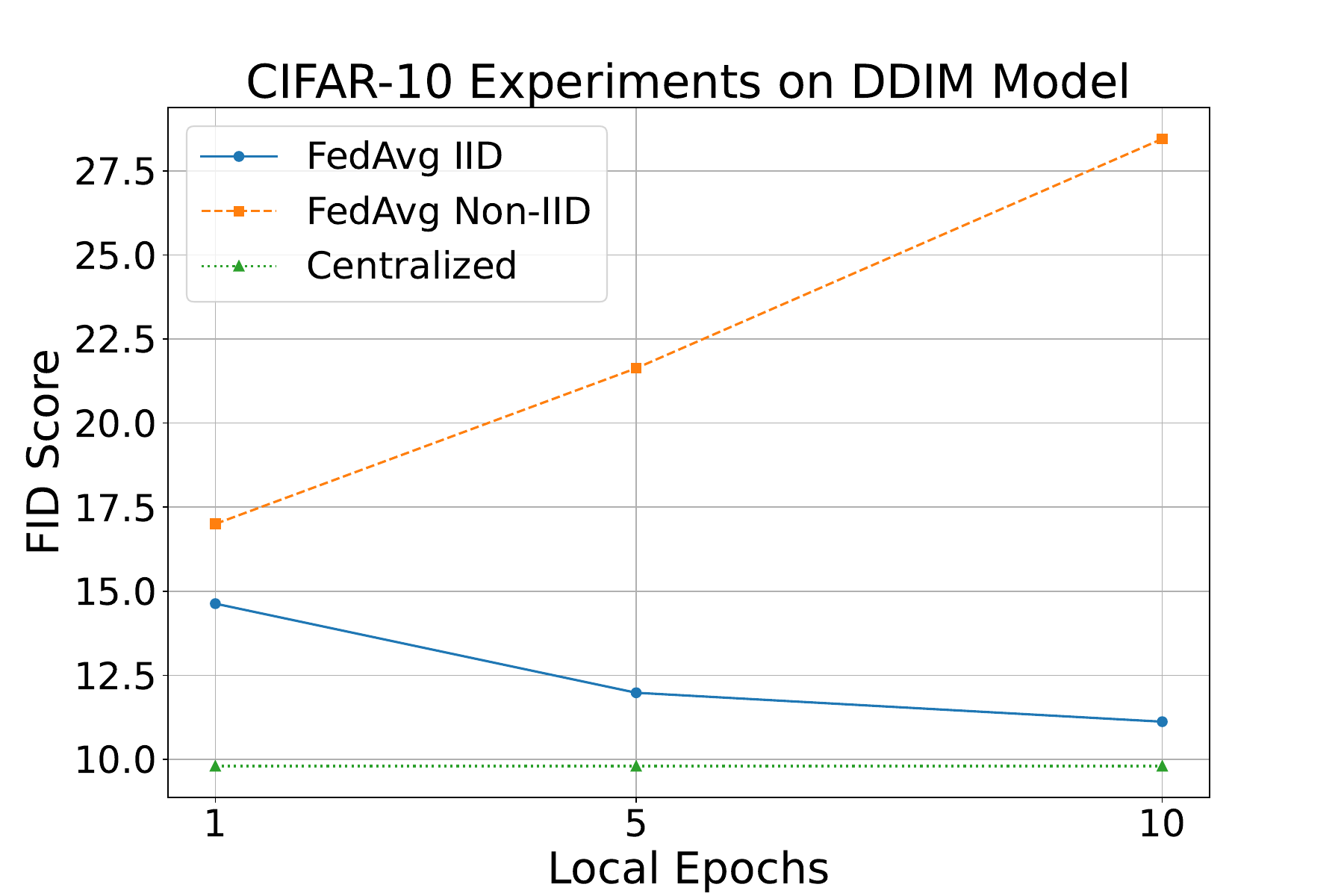}
    \caption{FID score comparison over CIFAR-10 data using DDIM \cite{Song2021} via FedAvg FL infrastructure and centralized training with IID and Non-IID data (number of clients is $K=20$, selection ratio is $\kappa=0.2$; more details are provided in Section \ref{sec:setting}).}
    \label{fig:cifar10_ddpm_fedavg}
\end{figure}
Recent efforts focus on leveraging DMs to solve challenges in FL, however, disregarding the reverse side, i.e., the issues when training DM in FL environments. Although training DMs is an unsupervised learning task, it faces challenges in FL, such as data heterogeneity, i.e., non-IID data \cite{StanleyJothiraj2024,Goede2024,Tun2023}, due to e.g., the bias and divergence of gradients in traditional supervised learning task \cite{Zhao2018,Li2020,Wang2020}. 
The performance degradation caused by such divergence 
is shown in Fig. \ref{fig:cifar10_ddpm_fedavg}.
Specifically, consider the Fréchet Inception Distance (FID) score, which assesses the quality of images created by a generative model; lower scores indicate that the generated images are more similar to real images w.r.t. quality. 
Fig. \ref{fig:cifar10_ddpm_fedavg} shows the FID score over the CIFAR-10 dataset using the Denoising Diffusion Implicit Model (DDIM \cite{Song2021}) in a FL environment adopting FedAvg \cite{McMahan2017}. 
One can observe the FID scores across different training epochs for FedAvg with IID data (independent and identically distributed) 
and non-IID data compared against centralized training. 
FedAvg using IID and non-IID data shows higher FID scores than the centralized approach, with non-IID data exhibiting significantly higher scores due to statistical heterogeneity. 

Limited research copes with the challenges of training DMs in FL environments. The Phoenix mechanism in \cite{StanleyJothiraj2024} aims to solve the non-IID data challenge by data sharing, which is not applicable in distributed learning environments. 
The authors in \cite{Goede2024} focus on training subsets of ML models while `freezing' the rest thus achieving communication efficiency in distributed learning. The approach in \cite{Tun2023} investigates the potential of training DMs for privacy-sensitive vision tasks. 
In this context, a research question arises: \textit{How can we obtain DMs that can generate relatively high-quality images efficiently trained in distributed learning environments?}

The results in Fig. \ref{fig:cifar10_ddpm_fedavg} and \cite{Deng2024} suggest that more frequent model aggregation mitigates the weight divergence problem over non-IID data. 
Therefore, we introduce FedPhD, a novel framework that addresses data heterogeneity, excessive communication, and computation costs through Hierarchical Federated Learning (HFL) and distributed structured pruning. The FedPhD framework
employs HFL by introducing the layer, \textit{the edge server layer}, to the traditional Cloud server-based FL framework. 
This additional layer engages edge servers between edge devices (clients) and the central server. Due to the proximity of edge servers to clients, frequent aggregation at the edge servers becomes feasible. To further reduce computation and communication costs, 
FedPhD leverages structured pruning for the U-Net (a convolutional neural network introduced for image segmentation) 
used in the diffusion process. Depending on devices' requirements and purposes, the structured pruning can be initially randomized or based on the \textit{$L_{2}$ group-norm} after the initial sparse training stage. Moreover, FedPhD introduces an edge-server selection mechanism in HFL to ensure data diversity and discrepancy-aware aggregation. The major contributions of this paper are:

\begin{itemize}
\item We demonstrate that frequent model aggregation addresses data heterogeneity in FL of DMs. This led to the introduction of our hierarchical FL paradigm, FedPhD.
\item We introduce a new edge-server selection and model aggregation strategy to ensure data homogeneity alleviating the non-IID data problem in HFL.
\item We integrate a structured pruning method for DMs into the HFL framework, significantly reducing communication and computation costs.
\item FedPhD achieves the best Fréchet Inception Distance (FID) and Inception Scores (IS) with $44\%$ less communication, computation, and storage requirements compared to well-known baselines \cite{McMahan2017}, \cite{Li2020a}, \cite{Goede2024}, \cite{li2021model}, \cite{karimireddy2020scaffold}. 
\end{itemize}


\section{Related Work}
\subsection{Training Diffusion Models in FL}
A few studies consider training DMs within FL environments. 
Those that achieve distributed DM training often provide a basic framework without addressing the key challenges of non-IID data and efficiency. \cite{yu2023federated} compared the advantages and disadvantages between federated Foundation Models (FM) and centralized FM optimization, generalizing the FM of collaborative learning. \cite{Goede2024} customize the FedDiffuse method into USPLIT, ULATDEC, and UDEC within the concept of split learning, which trains DMs with FedAvg \cite{McMahan2017} to split or limit the updates of parameters. This approach involves a central server collecting only parts of the DMs, thus reducing communication costs. 
Phoenix in \cite{StanleyJothiraj2024} is a DM learning framework designed to address non-IID data using a data-sharing policy and layer-wise personalization.

\subsection{Diffusion Models Compression}
DMs, though with powerful generative capabilities for high-quality images, are computationally intensive and memory-demanding, raising notable obstacles for their deployments on edge devices and clients. 
To address these concerns, various strategies have been proposed. 
Diff-Pruning \cite{Fang2024} is a pioneering method designed specifically for pruning DMs. 
This approach relies on Taylor expansion across pruned time steps acknowledging that not all time steps are equally important in DMs. 
\cite{gu2022vector} introduced VQ-Diffusion, a vector-quantized diffusion method based on a 
Vector-Quantized Variational Autoencoder (VQ-VAE) for text-to-image generation. \cite{shang2023post} proposed a 8-bit post-training quantization technique to accelerate the DM generation process while accounting for normally distributed time step calibration. 
\cite{huang2024knowledge} observed that smaller student models introduce more noise during the DM forward process, and thus proposed DiffKD. DiffKD is a knowledge distillation method where a trained teacher model denoises the student model’s features. 
To the best of our knowledge, FedPhD is the first method for applying pruning within the FL framework to train DMs in a hierarchical fashion. 

\section{Problem Fundamentals}
\label{sec:problem}
\subsection{Background on Diffusion Models}
Diffusion Models (DMs) are latent variable models inspired by non-equilibrium thermodynamics. 
\cite{Ho2020} proposed the Denoising Diffusion Probabilistic Model (DDPM), which leverages the \textit{diffusion process} and \textit{reverse process} to mimic the image generation process. 
The model progressively adds Gaussian noise to an image in the forward process and then iteratively denoises the image in the reverse process to recover the image from white noise. 
Assuming that data follow a distribution $q(\mathbf{x})$, 
a DM builds a generative distribution $p_{\theta}(\mathbf{x})$ with parameter $\theta$, that approximates the true distribution $q(\mathbf{x})$. This problem is formalized as estimating: 
\begin{equation}
\label{eq:reverse_process_untractable}
    p_{\theta}(\mathbf{x}) = \int p_{\theta}(\mathbf{x}_{0:T}) \, d\mathbf{x}_{1:T}, 
\end{equation}
where
\begin{eqnarray}
    p_{\theta}(\mathbf{x}_{0:T}) := p(\mathbf{x}_{T}) \prod_{t=1}^{T} p_{\theta}(\mathbf{x}_{t-1} \mid \mathbf{x}_{t})
\end{eqnarray}
is referred to as the \textit{reverse process}, which denoises the data departing from final data $\mathbf{x}_{T}$ to initial data $\mathbf{x}_{0}$ and $t$ represents the time step in the forward and reverse diffusion processes. 
The latent variables $\mathbf{x}_{1}, \ldots, \mathbf{x}_{T}$ have 
the same dimensionality as $\mathbf{x}_{0}\sim q(\mathbf{x}_{0})$. 
The Gaussian transition has the form: 
\begin{eqnarray}
p_{\theta}(\mathbf{x}_{t-1} \mid \mathbf{x}_{t}) = \mathcal{N}(\mathbf{x}_{t-1}; \mu_{\theta}(\mathbf{x}_{t}, t), \Sigma_{\theta}(\mathbf{x}_{t}, t)),
\end{eqnarray}
with white noise starting point at $p(\mathbf{x}_{T})=\mathcal{N}(\mathbf{x}_{T}; \mathbf{0},\mathbf{I})$. According to \cite{Ho2020}, Eq. \eqref{eq:reverse_process_untractable} is tractable conditioned on $\mathbf{x}_{0}$:
\begin{equation}\label{eq:reverse_process_tractable}
    q(\mathbf{x}_{t-1} \mid \mathbf{x}_{t}, \mathbf{x}_{0}) = \mathcal{N}(\mathbf{x}_{t-1}; \tilde{\boldsymbol{\mu}}_{t}(\mathbf{x}_{t}, \mathbf{x}_{0}), \tilde{\beta}_{t} \mathbf{I}),
\end{equation}
where
$
    \tilde{\boldsymbol{\mu}}_{t}(\mathbf{x}_{t}, \mathbf{x}_{0}) := \frac{\sqrt{\bar{\alpha}_{t-1}} \beta_{t}}{1 - \bar{\alpha}_{t}} \mathbf{x}_{0} + \frac{\sqrt{\alpha_{t}} (1 - \bar{\alpha}_{t-1})}{1 - \bar{\alpha}_{t}} \mathbf{x}_{t}
$
and $
    \tilde{\beta}_{t} := \frac{1 - \bar{\alpha}_{t-1}}{1 - \bar{\alpha}_{t}} \beta_{t}.
$
To approximate the posterior $q(\mathbf{x}_{1:T}\mid \mathbf{x}_{0})$, the \textit{forward process} is to model a Markov chain gradually adding Gaussian noise to the data, i.e.,
\begin{equation}\label{eq:forward_process}
    q(\mathbf{x}_{1:T} \mid \mathbf{x}_{0}) := \prod_{t=1}^{T} q(\mathbf{x}_{t} \mid \mathbf{x}_{t-1}),
\end{equation}
with
$q(\mathbf{x}_{t} \mid \mathbf{x}_{t-1}) := \mathcal{N}(\mathbf{x}_{t}; \sqrt{1 - \beta_{t}} \mathbf{x}_{t-1}, \beta_{t} \mathbf{I})$. The variance parameters $\beta_{t}$ are learnable but fixed in DDPM. With using the re-parameterization trick \cite{Ho2020}, 
the noise predictor is trained to optimize $\mathcal{L}(\theta)$, defined as 
\begin{equation}
\label{eq:noise_predict}
     \mathcal{L}(\theta) = \mathbb{E}_{t,\mathbf{x}_{0}, \boldsymbol{\epsilon}} \left[ \lVert \boldsymbol{\epsilon} - \boldsymbol{\epsilon}_{\theta}(\sqrt{\bar{\alpha}_{t}} \mathbf{x}_{0} + \sqrt{1 - \bar{\alpha}_{t}} \boldsymbol{\epsilon}, t) \rVert^{2} \right],
\end{equation}
where $\mathbf{x}_{0} \sim q(\mathbf{x}_{0})$, $\boldsymbol{\epsilon} \sim \mathcal{N}(\mathbf{0}, \mathbf{I})$, $\alpha_{t} = 1 - \beta_{t}$ and $\bar{\alpha}_{t} = \prod_{s=1}^{t} \alpha_{s}$.
After the simplified training process, DDPM generates the synthetic data through the iterative process, i.e., 
\begin{equation}\label{eq:ddpm_generate}
    \mathbf{x}_{t-1} = \frac{1}{\sqrt{\alpha_{t}}} \left( \mathbf{x}_{t} - \frac{\beta_{t}}{\sqrt{1 - \bar{\alpha}_{t}}} \boldsymbol{\epsilon}_{\theta}(\mathbf{x}_{t}, t) \right) + \sigma_{t} \mathbf{z},
\end{equation}
where $\mathbf{x}_{T}$ and $\mathbf{z}$ are the white noises following $\mathcal{N}(\mathbf{0},\mathbf{1})$. To balance the quality of the generative image and the inference efficiency, DDIM \cite{Song2021} generalizes \eqref{eq:ddpm_generate} as follows:
\begin{equation}\label{eq:ddim_generate}
\begin{aligned}
    \mathbf{x}_{t-1} = & \sqrt{\alpha_{t-1}} 
    \left( \frac{\mathbf{x}_{t} - \sqrt{1 - \alpha_{t}} \boldsymbol{\epsilon}_{\theta}(\mathbf{x}_{t})}{\sqrt{\alpha_{t}}} \right) \\
    & + \sqrt{1 - \alpha_{t-1} - \sigma_{t}^{2}} \cdot \boldsymbol{\epsilon}_{\theta}(\mathbf{x}_{t}) + \sigma_{t} \mathbf{z}
\end{aligned}
\end{equation}
 When $\sigma_{t}$ is defined as follows for all $t$:
\begin{equation}
    \sigma_{t} = \sqrt{\frac{(1 - \alpha_{t-1})(1 - \alpha_{t})}{1 - \alpha_{t}}} \sqrt{\frac{1 - \alpha_{t}}{\alpha_{t-1}}},
\end{equation}
then the forward process becomes a Markov chain, and the generative process is reduced to DDPM. 

The forward process of DDIM is non-Markovian, yet it does not require retraining the noise predictor. This is because in contrast of DDPM which uses a stochastic reverse process, DDIM employs a deterministic reverse process, which ensures that no additional noise is introduced at each step. Since DDIM does not introduce new noise during the reverse process, the time step \( t \) does not explicitly appear in \( \boldsymbol{\epsilon}_{\theta}(\mathbf{x}_{t}) \). 
The term \( \boldsymbol{\epsilon}_{\theta}(\mathbf{x}_{t}) \) in DDIM refers to the noise prediction for the current step, without any time dependency. 

\subsection{Hierarchical Federated Learning with DMs}
Without loss of generality, we consider a distributed learning system consisting of one server, a set of $\mathcal{N}_{e}$ edge servers, and a set of $\mathcal{N}$ distributed clients (edge devices), 
where $\lvert \mathcal{N}_{e} \rvert = N_{e} \ll N = \lvert \mathcal{N} \rvert$. 
In each communication round, $C=\kappa N$ clients are selected for aggregation in a typical FL system, like FedAvg, where $C \ll N$ and $\kappa \in (0,1)$ is the participation ratio. 
Each client, indexed by $n$, holds local data $\mathcal{D}_{n} = \{(\mathbf{x}_{i}^{(n)}, y_{i}^{(n)})\}$, where $\mathbf{x}_{i}^{(n)} \in \mathbb{R}^{d}$ and $y_{i}^{(n)}$ is the label. The total number of samples in client $n$ is $D_{n} = \lvert \mathcal{D}_{n} \rvert$.

The overall objective is to minimize the global loss function $F(\boldsymbol{\theta})$, which is a weighted average of the local loss functions $F_{n}(\boldsymbol{\theta})$ from each client $n$. In a FL system, the objective is:
\begin{equation}\label{eq:typical_fl}
    \min_{\boldsymbol{\theta}} F(\boldsymbol{\theta}) = \min_{\boldsymbol{\theta}} \sum_{n=1}^{N} \rho_{n} F_{n}(\boldsymbol{\theta}),
\end{equation}
where $\boldsymbol{\theta}$ denotes the global model parameters and $\rho_{n}=\frac{D_{n}}{D}$, $D = \sum_{n=1}^{N} D_{n}$ denotes the total number of samples across all clients. $F_{n}(\boldsymbol{\theta})$ is the local loss function for client $n$:
    \begin{equation}\label{eq:normal_loss}
        F_{n}(\boldsymbol{\theta}) = \frac{1}{D_{n}} \sum_{(\mathbf{x}_{i}^{(n)}, y_{i}^{(n)}) \in \mathcal{D}_{n}} \ell(\boldsymbol{\theta}; \mathbf{x}_{i}^{(n)}, y_{i}^{(n)}).
    \end{equation}
where $\ell(\boldsymbol{\theta}; \mathbf{x}, y)$ is the loss function for a single sample. In the task of training a DM, the local loss function \eqref{eq:normal_loss} becomes:
    \begin{equation}\label{eq:dm_loss}
        F_{n}(\boldsymbol{\theta}) = \frac{1}{D_{n}} \sum_{(\mathbf{x}_{i}^{(n)}) \in \mathcal{D}_{n}} \mathcal{L}(\boldsymbol{\theta}; \mathbf{x}_{i}^{(n)}),
    \end{equation}
where $\mathcal{L}(\boldsymbol{\theta}; \mathbf{x}_{i}^{(n)})$ follows \eqref{eq:noise_predict}. 
The process of training a DM within FL is provided in Algorithm \ref{alg:client_training_dm} (See Appendix ). 
To avoid stragglers, e.g., low computational capacity clients, \eqref{eq:typical_fl} reduces to
\begin{equation}\label{eq:typical_fl_select}
    \min_{\boldsymbol{\theta}} F(\boldsymbol{\theta}) = \min_{\boldsymbol{\theta}} \sum_{c=1}^{C} \frac{D_{c}}{D} F_{c}(\boldsymbol{\theta}).
\end{equation}

In HFL, $C$ clients conduct model aggregation on $\mathcal{N}_{e}$ edge servers rather than the central server with objective: 
\begin{equation}\label{eq:hfl_objective}
    \min_{\boldsymbol{\theta}} F(\boldsymbol{\theta}) = \min_{\boldsymbol{\theta}} \sum_{e=1}^{N_{e}} \sum_{n=1}^{M_{e}} \rho_{en} F_{en}(\boldsymbol{\theta}),
\end{equation}
where $M_{e}$ denotes the number of clients selected for edge server $e$ and $C= \sum_{e=1}^{N_{e}} M_{e} $. 
Regardless of the client selection and aggregation, the local training follows Stochastic Descent Gradients (SGD) updates. 
In HFL, during each communication round, the clients assigned to each edge server $e$ perform local updates and send their model updates to their respective edge servers. Each edge server $e$, in turn, aggregates these updates and sends the aggregated update to the central server. The global model update rule in HFL is:
\begin{equation}
    \boldsymbol{\theta}^{(r+1)} = \boldsymbol{\theta}^{(r)} - \eta \sum_{e=1}^{N_{e}} \sum_{n=1}^{M_{e}} \rho_{en} \nabla F_{en}(\boldsymbol{\theta}^{(r)}),
\end{equation}
where $r > 0$ is the communication round and $\nabla F_{en}(\boldsymbol{\theta}^{(r)})$ is the gradient of the local loss function for client $n$ on edge server $e$ with respect to the global model parameters.

\section{The FedPhD Framework}
\begin{figure*}[ht]
    \centering
    \includegraphics[width=0.7\textwidth]{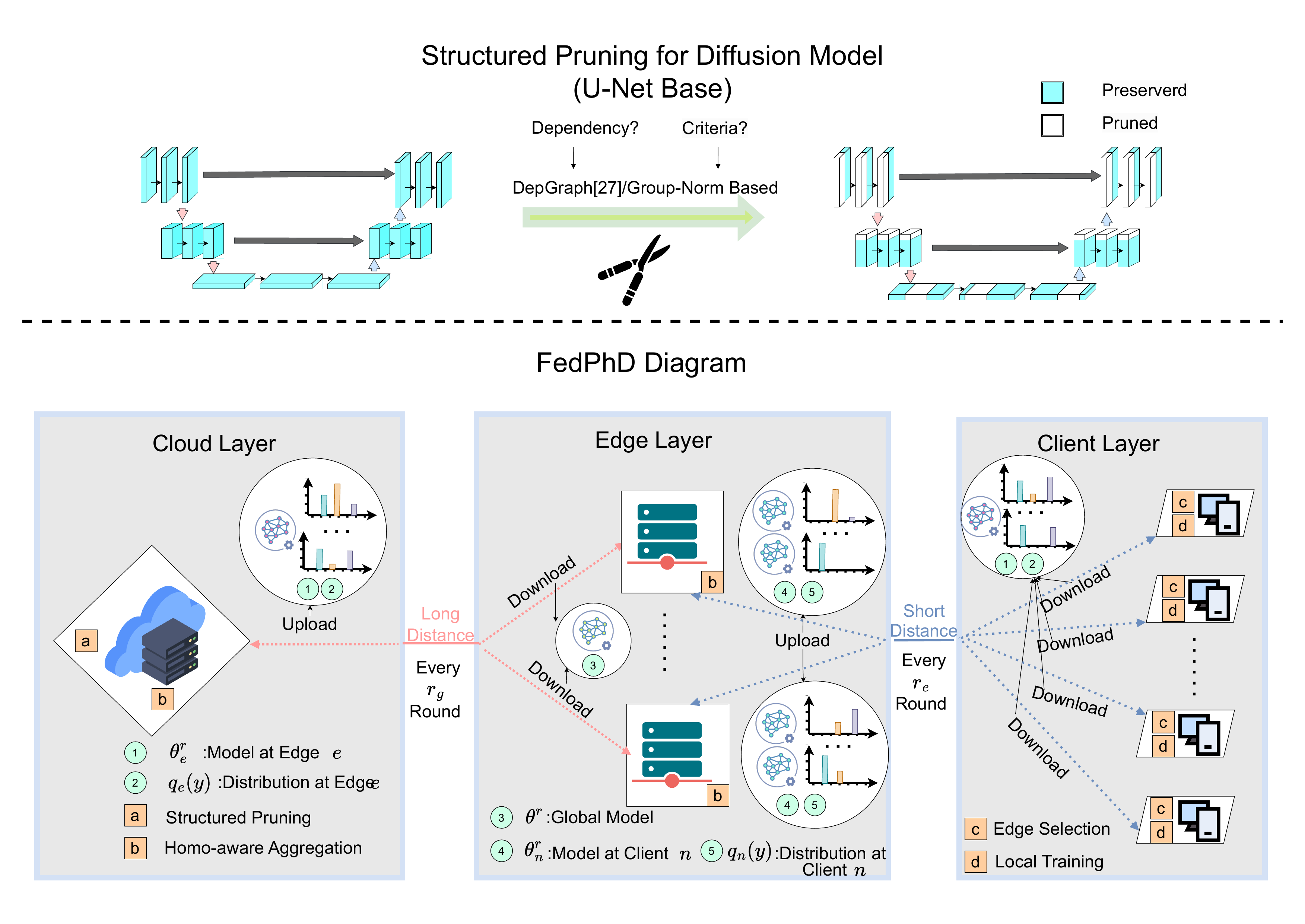}
    \caption{\textbf{Overview of FedPhD}: (Top) Illustration of structured pruning of a U-Net-based model, where DepGraph identifies dependencies, and group-norm criteria determine the pruned components; (bottom) Hierarchical FL (HFL) setup across Client, Edge, and Cloud layers. Homogeneity-aware aggregation is achieved on both: Cloud server and Edge servers to improve model consistency across diverse environments. 
    Structured pruning occurs only on the Cloud server after $R_{s}$ sparse training rounds (or at initialization). Clients perform local training and select Edge servers establishing client-edge associations $n \in \mathcal{M}_{e}$. Every $r_{g}$ round, active Edge servers connect to Cloud for central aggregation, uploading accumulated distributions $q_{e}(y)$ and models $\mathbf{\theta}_{e}^{r}$; while every $r_{e}$ round, selected clients upload updates $\mathbf{\theta}_{n}^{r}$ and client distribution $q_{n}(y)$ to their assigned Edge servers.}
    \label{fig:fedphd_overall}
\end{figure*}

\label{sec:fedphd}
FedPhD aims to efficiently train high-performance DMs in a HFL framework. As stated in ShapeFL \cite{Deng2024}, imbalanced data cause instability and divergence in HFL-based training, 
which advocates maximizing the \textit{data diversity} in aggregation. 
Hence, considering the computational and communication efficiency in the edge nodes, FedPhD leverages \textit{structured pruning} rather than unstructured pruning to achieve efficient compression. Subsequently, we introduce a \textit{homogeneity-aware} aggregation and edge server \textit{selection mechanism} to ensure data homogeneity in HFL. 
In the remainder of this section, we elaborate on the two main mechanisms of FedPhD.
\subsection{Structured Pruning for DMs}
Structured pruning removes entire groups, such as channels or layers, ensuring better CPU/GPU compatibility and more efficient model deployment than unstructured pruning.
DepGraph \cite{Fang2023} identifies \textit{layer dependencies} to guide structured pruning, making it adaptable to any model architecture. Diff-Pruning \cite{Fang2024} builds on DepGraph but assumes pre-trained DMs, limiting its applicability in FL. FedPhD overcomes this by integrating dependency-aware pruning during decentralized training. DepGraph models dependencies between adjacent layers in a symmetric matrix $\mathbf{S}$, capturing \textit{inter-layer} and \textit{intra-layer} relationships. This informs the parameter group $\mathcal{G}=\{\mathbf{\theta}^{1},\mathbf{\theta}^{2},\cdots, \mathbf{\theta}^{\lvert G \rvert}\}$ for structured pruning. 

FedPhD employs two pruning strategies based on hardware constraints:(1)\textbf{Pruning from scratch (FedPhD (OS)):} One-Shot (OS) pruning before training using an $L_2$-norm-based criterion for \textit{random pruning}.  (2)\textbf{Pruning after sparse training:} Group norm-based pruning applied after initial training rounds. 




For resource-limited clients, FedPhD-OS prunes before training, while clients with more capacity benefit from structured pruning after initial training. The local sparse training loss at client \(n\) is:
\begin{equation}
\label{eq:dm_loss_sparse}
    F_{n}(\boldsymbol{\theta}) = \frac{1}{D_{n}} \sum_{(\mathbf{x}_{i}^{(n)}) \in \mathcal{D}_{n}} \mathcal{L}(\boldsymbol{\theta}; \mathbf{x}_{i}^{(n)}) + \mathbf{\Omega}(\mathcal{G}, k),
\end{equation}
where \(\mathbf{\Omega}(\mathcal{G}, k)\) denotes a regularization term defined as:
\begin{equation}
\label{eq:group_norm}
    \mathbf{\Omega}(\mathcal{G}, k) = \sum_{k=1}^{K}\sum_{\mathbf{\theta}^{g} \in \mathcal{G}} \lambda_{g} \|\mathbf{\theta}^{g}[k]\|^{2}_{2}.
\end{equation}
The \(\lambda_{g}\) factor refers to the group regularization scale parameter, while 
\(\mathbf{\theta}^{g}[k]\) denotes the \(g^{\text{th}}\) group of independent prunable parameters 
at the \(k^{\text{th}}\) prunable dimension.
Considering the specific U-Net architecture, the redundancy in the middle layers is greater than that in input and output layers. Hence, we introduce a mechanism to determine \(\lambda_{g}\). 
First, we calculate the mean layer index of \(\mathbf{\theta}\) as \(l_{\text{mid}}\). For each layer \(l\) in group \(\mathbf{\theta}^{g}\), the distance between the mean layer and \(l\) is given by \(\lvert l - l_{\text{med}} \rvert\). The score for the entire group \(\mathbf{\theta}^{g}\) is calculated by:
    $Q(\mathbf{\theta}^{g}) = \frac{1}{L} \sum_{l \in \mathbf{\theta}^{g}} \lvert l - l_{\text{med}} \rvert,$
where \(L\) denotes the number of layers in group \(\mathbf{\theta}^{g}\). 
Therefore, the parameter \(\lambda_{g}\) for group \(\mathbf{\theta}^{g}\) is defined as inversely proportional to score \(Q(\mathbf{\theta}^{g})\), i.e., $    \lambda_{g} = \frac{\lambda_0}{Q(\mathbf{\theta}^{g})}$
where \(\lambda_0\) is a regularization factor.

\subsection{Homogeneity-Aware Aggregation and Server Selection}
We propose a novel aggregation method for HFL that addresses data imbalance by leveraging a Statistical Homogeneity (SH) score and an edge selection policy. Specifically, we introduce the concept of homogeneity-aware aggregation, detail the computation of the SH score, and illustrate how it guides both client selection and model aggregation.
ShapeFL \cite{Deng2024} measures \textit{data diversity} using cosine distance between model updates but is computationally expensive and requires manual layer selection, limiting its applicability. Additionally, its edge server selection relies on offline pre-training, which is unsuitable for FL. Instead, we propose a lightweight approach that focuses on data distribution rather than model-specific layers.
Following \cite{Li2021, Ye2023}, the \textit{uniform} distribution serves as a reference point for evaluating statistical data homogeneity. Given local data $\mathcal{D}_n = {(\mathbf{x}_i, y_i)}$ on client $n$ with label distribution $q_n$, we define $q_u$ as the uniform distribution over target labels $\mathcal{Y}$. If the global distribution is non-uniform, $q_u$ is set as target label distribution accordingly. The SH score $\mu_n$ for client $n$ is:
\begin{equation}
\label{eq:stat_homo}
\mu_n = 2 - \sqrt{\sum_{y \in \mathcal{Y}} \left| q_n(y) - q_u(y) \right|^2},
\end{equation}
where $q_n(y)$ is the frequency of label $y$ in client's dataset and $q_u(y) = \frac{1}{|\mathcal{Y}|}$ is the uniform probability of $y$.

\begin{algorithm}[h]
\caption{The FedPhD Algorithm for HFL of DMs}
\label{alg:fedphd}
\begin{algorithmic}[1]
\STATE \textbf{Input:} Initial global model $\boldsymbol{\theta}^{(0)}$; global and sparse training rounds $R$, $R_{s}$; edge and global update frequency, $r_e$, $r_g$; number of edge servers and clients $N_e$, $N$; SH score parameters $a$ and $b$; target distribution $q_u(y)$; pruning ratio $s_{p}$
\STATE Initialize $q_e(y)$ and  $\{n_e\} = \{0\}$ for all $e \in \mathcal{N}_e$
\FOR{each global round $r = 1, \dots, R$}
    \FOR{each client $n \in \mathcal{C}$}
        \STATE Select edge server $e$ with probability $P_{n}(e)$ in \eqref{eq:edge_server_selection}
    \ENDFOR
    \FOR{each edge server $e \in \mathcal{N}_e$ in parallel}
        \STATE Broadcast  $q_e(y)$ to all clients
        \FOR{each client $n \in \mathcal{M}_{e}$ in parallel}
            \IF{$r < R_{s}$}
                \STATE Perform Algorithm \ref{alg:client_training_dm} on \eqref{eq:dm_loss_sparse}
            \ELSE
                \STATE Perform Algorithm \ref{alg:client_training_dm} on \eqref{eq:dm_loss}
            \ENDIF
            \STATE Send updated model $\boldsymbol{\theta}_{n}^{r+1}$ to edge server $e$
        \ENDFOR
        \STATE Update $n_e \leftarrow n_e + \sum_{n \in \mathcal{M}_{e}}n_n$
        \IF{$r \mod r_{e} = 0$}
                \STATE Update $q_e(y)$ with \eqref{eq:update_edge_distribution} 
                \STATE Aggregation using Eqs. \eqref{eq:stat_homo_hfl}, \eqref{eq:agg_server_model}, and \eqref{eq:weight_client}
                \STATE Send back $\boldsymbol{\theta}^{r+1}_{e}$ to client $n \in \mathcal{M}_{e}$ 
            \ENDIF
    \ENDFOR
    
    \IF{$r \mod r_g = 0$}
        \STATE Request $q_e(y)$ from edge servers.
        \STATE Aggregation using Eqs. \eqref{eq:agg_global_model}, \eqref{eq:weight_edge_server}, and \eqref{eq:stat_homo_hfl}
        \IF{$r = R_{s}$}
            \STATE  Structured pruning on $\boldsymbol{\theta}^{(r+1)}$ with $s_{p}$
        \ENDIF
        \STATE Send back $\boldsymbol{\theta}^{r+1}$ to all clients.
        \STATE Re-initialize $q_e(y)$ and  $\{n_e\} = \{0\}$ for all $e \in \mathcal{N}_e$
    \ENDIF
\ENDFOR
\STATE \textbf{Output:} Final global model parameters $\boldsymbol{\theta}^{(R)}$
\end{algorithmic}
\end{algorithm}
According to \cite{Li2021} and \cite{Zhao2018}, SH is closely related to data diversity, as it reflects the extent to which the global model approximates the performance of a model trained on IID data. In this context, we aim for a DM that generalizes well on non-IID data by using techniques that maximize SH. However, in HFL, clients are assigned directly to edge servers for aggregation, which makes the SH score at \textit{client} level in \eqref{eq:stat_homo} insufficient. 
Therefore, we compute the SH score at \textit{edge server} level for the aggregation on the central server. 
In HFL, the \textit{accumulated distribution} for an edge server is the weighted sum of the label distributions from all clients assigned to that edge server. 
This accumulated distribution must be \textit{refreshed} (re-initialized) dynamically at a specific frequency to maintain an accurate representation of the data of the participating clients. In HFL, aggregation occurs more frequently than in FL, so this refreshment is synchronized with the frequency of FL aggregation, which is denoted by $r_{g}$ in this work.
Specifically, given a current accumulated distribution $q_{e}(y)$ and attached client samples $n_{e}$,
if each client $n$ contributes \( n_n \) samples to the edge server then the updated accumulated distribution \( q_{e}^{\prime} \) for edge server $e$ is updated as:
\begin{equation}\label{eq:update_edge_distribution}
    q_{e}^{\prime}(y) = \frac{q_{e}(y)+\sum_{n \in \mathcal{M}_e} q_n(y) \cdot n_n}{n_{e}+\sum_{n \in \mathcal{M}_e} n_n},
\end{equation}
where \( \mathcal{M}_e \) is the set of clients assigned to edge server $e$ and \( q_n(y) \) is the proportion of label \( y \) in client $n$’s dataset. 
By continually updating the accumulated distribution, the edge server maintains an accurate representation of its clients' data diversity. This ensures that the subsequent model aggregation steps consider the underlying data distribution of participated clients. 
Consequently, the SH score for edge server $e$ in HFL is similarly calculated as in \eqref{eq:stat_homo} by using the accumulated distribution $q_{e}(y)$ in \eqref{eq:update_edge_distribution}, which is computed as:
\begin{equation}\label{eq:stat_homo_hfl}
    \mu_{e} = 2 - \sqrt{\sum_{y \in \mathcal{Y}} \left| q_e(y_e = y) - q_u(y_u = y) \right|^2},
\end{equation}
where $q_{u}(y)$ represents the target (uniform) distribution at the central server.
Note that neither the central server nor the edge servers possess any raw data.

Based on SH, we develop a strategy for homogeneity-aware aggregation and edge server selection in HFL. The aggregation process takes place at both the central server and edge server levels. Starting at the central server level, each edge server's contribution to the global model is weighted by a factor that considers both the number of samples it has aggregated and its SH. 
Let $\mathbf{\theta}_{e}$ denote the model parameters from edge server $e$, and let $n_e$ represent the corresponding number of samples. The central server aggregates the global model parameters $\mathbf{\theta}$ using 
the weighted sum: 
\begin{equation}
\label{eq:agg_global_model}
    \mathbf{\theta} = \sum_{e=1}^{ N_{e}} \rho_e \cdot \mathbf{\theta}_{e},
\end{equation}
where \( \rho_e \) is the weight for edge server $e$, calculated as:
\begin{equation}\label{eq:weight_edge_server}
    \rho_e = \frac{\textit{ReLu}(n_e + a \cdot \mu_e + b)}{\sum_{e=1}^{N_{e}} \textit{ReLu}(n_e + a \cdot \mu_e + b)}.
\end{equation}
$\mu_e$ represents the SH of the edge server, $a$ and $b$ are tunable parameters that balance the contribution of the sample size and SH; \textit{ReLu}$(x) = \max(0,x)$.
This approach ensures that edge servers with higher SH, i.e., indicating a distribution closer to 
target distribution, contribute more significantly to the global model update, thereby improving the overall performance of the global model on non-IID data \cite{Ye2023}. 
Similarly, the edge server aggregates the edge server model $\theta_{e}$ according to:
\begin{equation}\label{eq:agg_server_model}
    \theta_{e} = \sum_{n\in \mathcal{M}_{e}}\rho_{en} \theta_{n},
\end{equation}
where \( \rho_{en} \) is the weight for selected clients for aggregation on edge server $e$, computed as:
\begin{equation}\label{eq:weight_client}
    \rho_{en} = \frac{\textit{ReLu}(n_n + a \cdot \mu_n + b)}{\sum_{n\in \mathcal{M}_{e}} \textit{ReLu}(n_n + a \cdot \mu_n + b)}
\end{equation}

Note that unlike traditional FL frameworks such as FedAvg, FedPhD allows clients to select their edge servers for aggregation. This selection process is guided by the SH, which evaluates how closely an edge server's accumulated distribution aligns with the target distribution. For each client $n$, the selection of the edge server $e$ is determined based on a probability distribution, expressed as: 
\begin{equation}\label{eq:edge_server_selection}
    P_{n}(e) = \frac{\textit{ReLu}\left( \mu_{e}^{n\prime} \cdot a - n_{e}^{n\prime} + b \right)}{\sum_{e=1}^{N_{e}} \textit{ReLu}\left( \mu_{e}^{n\prime} \cdot a - n_{e}^{n\prime} + b \right)},
\end{equation}
where \( n_{e}^{n\prime} \) and \( \mu_{e}^{n\prime} \) represent the updated sample size and SH score for edge server $e$ after including client $n$, respectively. The client is then assigned to the edge server $e$ based on the probability distribution \( P_{n}(e) \), ensuring a higher likelihood of selection for an edge server that better matches the target distribution while also considering the load balance by favoring servers with fewer clients communicated before. 

\subsection{The FedPhD Algorithm}
The workflow of FedPhD is illustrated in Algorithm \ref{alg:fedphd} and Fig. \ref{fig:fedphd_overall}. We outline the processes of clients, edge servers, and the central server in the FedPhD algorithm as follows. 

\subsubsection{Central Server Process} 
The homogeneity-aware aggregation and structured pruning occur in the central server. If constraints such as memory limit exist for devices, the central server performs \textit{random} pruning initially (i.e., at $r=0$); otherwise, the central server performs group-norm based pruning at $r=R_{s}$ after $R_{s}$ sparse training rounds (\textit{lines 24-26}). The central server performs aggregation every $r_{g}$ rounds iteratively, maximizing the SH score to counter the data heterogeneity issue in HFL (\textit{lines 22-23}). The model parameters are received from the edge servers, which are the aggregations of respective clients.

\subsubsection{Edge Server Process} 
Edge servers aggregate models more frequently than the central server ($r_{e}<r_{g}$, \textit{line 19}). After aggregation, the aggregated results are returned to their respective clients (\textit{line 20}). In addition, they maintain and update the accumulated distribution based on the model updates received from clients (\textit{line 18}).

\subsubsection{Client Process} 
The client is responsible for selecting the edge servers and conducting local training of DMs. 
At each round, the client selects the edge server for aggregation based on the probability $P_{n}(e)$ as described in \eqref{eq:edge_server_selection}, ensuring a higher SH for aggregation (\textit{lines 4-5}). If sparse training is required and in initial rounds, i.e., $r<R_{s}$, the group regularization term is added to the local loss function, and the client performs Algorithm \ref{alg:client_training_dm} on the loss function as in \eqref{eq:dm_loss_sparse}; otherwise, it uses the loss function as in \eqref{eq:dm_loss}. The procedure is shown in Algorithm \ref{alg:fedphd} (\textit{lines 8-12}).

Note, both the label distribution of clients and aggregated distribution of edge servers are necessary to compute SH. 
Edge servers track the client distributions, where each edge server $e$ holds the values $q_{n}(y)$ and $n_{n}$ for all clients $n$. 
As a result, the label distribution does not need to be communicated during the FedPhD training process. For the accumulated (updated) distribution, computed at the edge servers, $q_{e}(y)$ is transmitted to clients during each global round for edge server selection and to the central server every $r_{g}$ rounds for homogeneity-aware aggregation. Since only distribution information is shared, FedPhD ensures stronger privacy preservation compared to Phoenix \cite{StanleyJothiraj2024}, with negligible additional communication overhead.

\subsection{Balancing Efficiency and Privacy in FedPhD}
FedPhD uses information on the category distribution of client data to perform homogeneity-aware aggregation, which improves the model convergence and eases the data heterogeneity issue. However, sharing these label distributions could potentially expose sensitive information, raising privacy concerns. To mitigate these concerns, we introduce several privacy-preserving strategies that could be integrated into the FedPhD framework: (i) \textbf{Noise Injection}: Using differential privacy methods, such as adding controlled noise to category counts, results in individual data contributions that are less identifiable. (ii)  \textbf{Secure Aggregation}: Employing techniques like secure Multi-Party Computation (MPC) to combine distributions that prevent the central server from identifying individual client information. Such strategies are proposed to strike a balance between improving the model’s efficiency and protecting the privacy of client data. While these methods are not currently implemented in our FedPhD, they offer a pathway to enhance privacy in our future agenda.

\section{Experimental Evaluation}
\begin{table*}[h]
\centering
 \caption{Quality Metrics Evaluation on CIFAR10 and CelebA Datasets. 
 $A=10,000$ for CIFAR10 and $B=5,000$ for CelebA}
\begin{tabular}{lcccc}
\hline
\multirow{2}{*}{Method} & \multicolumn{2}{c}{CIFAR10} & \multicolumn{2}{c}{CelebA} \\
\cline{2-5}
& FID $\downarrow$ & IS $\uparrow$ & FID $\downarrow$ & IS $\uparrow$ \\
\hline
\textbf{Centralized and IID Settings} & & & & \\
Centralized Training & 8.73 $\pm$ 0.30 & 7.22 $\pm$ 0.21 & 5.86 $\pm$ 0.36 & 3.22 $\pm$ 0.14 \\
FedAvg (IID) & 11.98 $\pm$ 0.38 & 5.18 $\pm$ 0.13 & 6.42 $\pm$ 0.34 & 2.92 $\pm$ 0.22 \\
\hline
\textbf{Computation Equivalence (A/B Local Epochs)} & & & & \\
FedAvg & 24.50 $\pm$ 0.38 & 3.88 $\pm$ 0.12 & 14.89 $\pm$ 0.64 & 2.47 $\pm$ 0.15 \\
FedProx & 26.53 $\pm$ 0.82 & 3.71 $\pm$ 0.16 & 16.33 $\pm$ 0.34 & 2.45 $\pm$ 0.10 \\
FedDiffuse & 29.38 $\pm$ 0.52 & 3.56 $\pm$ 0.19 & 17.42 $\pm$ 0.58 & 2.32 $\pm$ 0.11 \\
MOON & 22.13 $\pm$ 0.66 & 4.11 $\pm$ 0.29 & 12.56 $\pm$ 0.31  & 2.44 $\pm$ 0.14 \\
SCAFFOLD & 43.87 $\pm$ 2.76 & 3.12 $\pm$ 0.33 & 32.32 $\pm$ 1.22  & 2.12 $\pm$ 0.28 \\
FedAvg ($E=1$) & 17.21 $\pm$ 0.28 & \textbf{4.68} $\pm$ 0.23 & 9.55 $\pm$ 0.32 & 2.78 $\pm$ 0.13 \\
FedPhD & \textbf{16.74} $\pm$ 0.26 & 4.24 $\pm$ 0.17 & 8.32 $\pm$ 0.44 & 2.76 $\pm$ 0.21 \\
FedPhD (OS) & 17.03 $\pm$ 0.22 & 4.14 $\pm$ 0.16 & \textbf{7.48} $\pm$ 0.28 & \textbf{2.88} $\pm$ 0.19 \\
\hline
\textbf{Fixed Communication Rounds (A/B Rounds)} & & & & \\
FedAvg & 21.63 $\pm$ 0.50 & 4.16 $\pm$ 0.12 & 11.29 $\pm$ 0.42 & 2.81 $\pm$ 0.20 \\
FedProx & 22.42 $\pm$ 0.42 & 4.01 $\pm$ 0.18 & 13.39 $\pm$ 0.38 & 2.55 $\pm$ 0.20 \\
FedDiffuse & 25.38 $\pm$ 0.54 & 3.66 $\pm$ 0.11 & 15.42 $\pm$ 0.58 & 2.34 $\pm$ 0.12 \\
MOON & 20.43 $\pm$ 0.63 & 4.17 $\pm$ 0.22 & 10.88 $\pm$ 0.42  & 2.78 $\pm$ 0.22 \\
SCAFFOLD & 29.54 $\pm$ 0.52 & 3.54 $\pm$ 0.22 & 21.44 $\pm$ 0.32 & 2.38 $\pm$ 0.23  \\
\hline
\end{tabular}
\label{tab:quality_metrics}
\end{table*}

\subsection{Experiment Setup}\label{sec:setting}
\subsubsection{Datasets and Models}
The effectiveness of FedPhD is validated using the CIFAR-10 and CelebA benchmarks with the DDIM model \cite{Song2021}, employing $100$ time steps, compared to the full $1000$ time steps used by DDPM \cite{Ho2020}. We focus exclusively on the training data, consisting of $50K$ images of size $32\times32$ in CIFAR10 and $163K$ images of size $64\times64$ in CelebA. We employ the same U-Net architecture as in \cite{Ho2020}, where the dense model comprises $35.7$ million parameters.

\subsubsection{Configurations}
All experiments are conducted using PyTorch on NVIDIA GeForce RTX 4090 and A6000 GPUs, with each experiment running on a single GPU. The experimental setup consists of $N=20$ clients, each connected to one of $N_{e}=2$ edge servers, with a central server overseeing the HFL process. 
Central aggregation occurs every $r_{g}=5$ rounds, while edge server aggregation is performed in every round, i.e., $r_{e}=1$. 
In CelebA, we partition the data based on two binary attributes, 
\textit{Male} and \textit{Young}, resulting in four classes: `young male', `old male', `young female', and `old female'. 
To devise non-IID data cases, each client is assigned images 
from only one class. In CIFAR10, each client holds images 
from $2$ classes. The label distributions across clients are illustrated in Figure \ref{fig:label_distributions}. All experiments are conducted with the same number of total local epochs, aligning with the same number of data sampling batches in the DM training. For the baseline methods, we set the number of local epochs per round to $E=5$, 
given that $r_{g}=5$, and the total number of communication rounds is $R=2000$. 
To ensure computational load equivalence, FedPhD is configured with $R=10000$ communication rounds and $E=1$ local epoch per round. 
In CelebA, the baselines are set similarly with $R=1000$ and $E=5$, while FedPhD is configured with $R=5000$ and $E=1$. Additionally, the experiments of fixed communication rounds are conducted for all baselines.
We use $500K$ training steps in the Centralized Training experiments and employ the Exponential Moving Average (EMA) for model updates as outlined in \cite{Ho2020}. 

\subsubsection{Baselines} 
The characteristics of baselines used for fair comparison are summarized in Table \ref{tab:baselines} and the details are included in Appendix.

\begin{table}[H]
\caption{Overall comparison of all methods based on computation, communication, and data heterogeneity.}
\centering
\begin{tabular}{|l|c|c|c|}
\hline
\textbf{Algorithm} & \textbf{Comp. Eff.} & \textbf{Comm. Eff.} & \textbf{Heterogeneity} \\
\hline
FedAvg \cite{McMahan2017}     & \ding{55} & \ding{55} & \ding{55} \\
FedProx \cite{Li2020a}   & \ding{55} & \ding{55} & \ding{51} \\
FedDiffuse \cite{Goede2024} & \ding{51} & \ding{51} & \ding{55} \\
MOON \cite{li2021model}   & \ding{55} & \ding{55} & \ding{51} \\
SCAFFOLD \cite{karimireddy2020scaffold}   & \ding{55} & \ding{55} & \ding{51} \\
FedPhD (\textbf{Ours})     & \ding{51} & \ding{51} & \ding{51} \\
\hline
\end{tabular}

\label{tab:baselines}
\end{table}

\subsubsection{Evaluation Metrics}
We evaluate all methods using two main types of metrics: 
efficiency and quality. \textbf{(\textit{i}) Quality Metrics:} To assess the quality of the generated images, we use the Fréchet Inception Distance (FID) and Inception Score (IS), as outlined in \cite{Ho2020,Song2021}. FID measures the distance between the distributions of synthetic images and real images, providing a quantitative measure of the realism of generated content. Additionally, we adopt the IS to evaluate the diversity of the generated data. 
All images are generated by DDIM \cite{Song2021} with steps $t=100$. 
The number of samples generated for measurements is $30K$ with batch size $256$. \textbf{(\textit{ii}) Efficiency Metrics:} Efficiency is measured by the number of parameters (\#Params), Multiply-Add Accumulation (MACs) and the communication volume per central aggregation in gigabytes (GB). 
The central aggregation process is defined as the period from one aggregation event to the next on the central server.
\begin{table*}[h]
\caption{Model pruning with different ratios for CIFAR-10 and CelebA datasets. The '-' symbol represents the percentage decrease in performance for quality metrics compared to the dense model ($0\%$ ratio).}
\centering
\begin{tabular}{|c|c|c|c|c|c|c|c|}
\hline
\textbf{Ratio} & \textbf{\#Params} & \textbf{MACs (CIFAR-10)} & \textbf{MACs (CelebA)} & \textbf{FID (CIFAR-10)} $\downarrow$ & \textbf{IS (CIFAR-10)} $\uparrow$ & \textbf{FID (CelebA)} $\downarrow$ & \textbf{IS (CelebA)} $\uparrow$ \\ \hline
0\%  & 35.7M & 6.1 & 24.3 & 16.56 & 4.42 & 7.31 & 2.92 \\ \hline
25\% & 26.9M & 4.2 & 16.8 & 16.78 (-1.3\%) & 4.18 (-5.4\%) & 7.42 (-1.5\%) & 2.84 (-2.7\%) \\ \hline
44\% & 20.3M & 3.4 & 13.6 & 16.74 (-1.1\%) & 4.24 (-4.1\%) & 7.48 (-2.3\%) & 2.88 (-1.4\%) \\ \hline
61\% & 13.9M & 2.1 & 8.3 & 19.36 (-16.9\%) & 3.83 (-13.3\%) & 9.59 (-31.2\%) & 2.73 (-6.5\%) \\ \hline
74\% & 9.3M  & 1.5 & 5.9 & 22.34 (-35.0\%) & 3.78 (-14.5\%) & 13.44 (-83.9\%) & 2.44 (-16.4\%) \\ \hline
\end{tabular}
\label{tab:pruning_ratios}
\end{table*}

\begin{table}[h]
\centering
\caption{Efficiency Metrics Evaluation on CIFAR10.}
\begin{tabular}{lccc}
\hline
Method & \#Params (M) $\downarrow$ & MACs (G) $\downarrow$ & Comm Cost $\downarrow$ \\
\hline
FedAvg & 35.7 & 6.06 & 109.22 \\
FedAvg ($E=1$) & 35.7 & 6.06 & 546.10 \\
FedProx & 35.7 & 6.13 & 109.22 \\
FedDiffuse & 35.7 & 6.06 & 72.82 \\
MOON      &  35.7  & 6.26 & 109.22  \\
SCAFFOLD & 35.7 & 6.13 & 218.44 \\
FedPhD & \textbf{20.3} & \textbf{3.42} & \textbf{65.46} \\
\hline
\end{tabular}
\label{tab:efficiency_metrics}
\end{table}
\subsection{Model Evaluation}
The FedPhD(OS) method refers to adopting \textit{One-Shot pruning} before training. The results presented in Table \ref{tab:quality_metrics} are averaged over the best outcomes across $A$ communication rounds and local epochs for CIFAR-10 ($A=10,000$) and $B$ communication rounds and local epochs for CelebA ($B=5,000$). Each setup was evaluated using at least three random seeds to ensure statistical reliability. 
From Table \ref{tab:quality_metrics}, we observe that despite not applying EMA updates in the distributed setting, FedAvg demonstrates reasonable performance under the IID data setup. However, it still underperforms compared to centralized training. When the data are non-IID, the quality of the images generated from FedAvg models degrades significantly, with the FID score increasing from $11.98$ to $21.63$ on CIFAR10, and from $6.42$ to $11.29$ on CelebA. 

More frequent aggregation by reducing $E$ from 5 to 1, helps mitigate the effects of data heterogeneity, albeit at the cost of a $5x$ increase in communication overhead. 
The well-known \textit{FedProx}, which balances communication and computation costs, also fails to address the data heterogeneity problem. As expected, \textit{FedDiffuse} performs worse than FedAvg, since it builds on the FedAvg framework while adopting the split learning paradigm. Therefore, it inherits the limitations of FedAvg's performance under heterogeneous data. SCAFFOLD underperforms compared to FedAvg in training DMs, which is due to client drift not being adequately corrected by the control variates method (due to the denoising process). Similar worse results of SCAFFOLD dealing with non-iid issues have been shown in \cite{li2021model}. MOON mitigates data heterogeneity to some extent by leveraging contrastive learning, resulting in a reduction in FID. 
In comparison, \textit{FedPhD} outperforms all baselines, achieving an FID score of 16.74 and an IS score of 4.24 on CIFAR10, and an FID score of 7.48 and IS score of 2.88 on CelebA. These results indicate that FedPhD successfully overcomes the degraded performance typically seen in distributed training under data heterogeneity, particularly when training DMs.

\subsection{Efficiency Evaluation}
As shown in Table \ref{tab:efficiency_metrics}, FedPhD reduces model parameters and MACs by $44\%$ on CIFAR10 with a pruning ratio of $44\%$. For CelebA, the primary difference lies in MACs, since the input image size is twice that of CIFAR10 while using the same model. Consequently, CelebA’s MACs are four times higher, with baseline models at $24.24$ and FedPhD at $13.68$.

To evaluate communication volume, we follow ShapeFL’s approach \cite{Deng2024}, where the communication cost depends on the physical distance between communicating clients and the transmitted data volume. Specifically: \( C_{ne} = 0.002 \cdot d_{e} \cdot V \), where \( d_{e} \) is the shortest path distance between node \( n \) and edge server \( e \), and \( V \) is the volume of transmitted data. The communication cost between the edge server and the central server is defined as \( C_{ce} = 0.02 \cdot d_{c} \cdot V \), given that $d_{c}$ is $10\times$ greater than $d_{e}$.
We provide the standardized communication volume in Table \ref{tab:efficiency_metrics}. 

With FedAvg model sizes at $136.53$ MB and FedPhD at $77.93$ MB (for $s_p = 44\%$), FedPhD reduces communication and computational costs by $44\%$ and $40\%$, respectively, while reducing memory requirements. Notably, FedPhD achieves a more compact model while outperforming baselines (see Table \ref{tab:quality_metrics}), making it ideal for deployment in resource-constrained edge environments.
Overall, FedPhD reduces communication costs by up to ~$88\%$ while maintaining or even improving performance.

\subsection{Discussion of different Ratio $s_{p}$}
Table \ref{tab:pruning_ratios} shows that pruning significantly reduces both model size and computational cost, with parameter and MACs reductions of up to $74\%$ and $75\%$, respectively. Up to a pruning ratio of $44\%$, FedPhD maintains stable performance, with minimal degradation in image quality, as reflected by small changes in FID (from $16.56$ to $16.74$) and IS (from $4.42$ to $4.24$). As the pruning ratio increases beyond $44\%$, there is a noticeable drop in performance, with FID increasing to $19.36$ at $61\%$ pruning and $22.34$ at $74\%$, while IS declines to $3.83$ and $3.78$, respectively. This approach leverages the strength of DepGraph, which efficiently captures inter-layer and intra-layer dependencies, along with the proposed group-norm pruning method. 
These results indicate that FedPhD tolerates moderate pruning (up to $44\%$) with minimal impact on image quality, while significantly reducing computational overhead. Notably, even with a pruning ratio of $44\%$, FedPhD still provides better quality images compared to baselines in non-IID settings, as shown in Table \ref{tab:quality_metrics}. On the other hand, more `aggressive' pruning leads to considerable performance degradation, suggesting a trade-off between efficiency and image generation quality.

\section{Conclusions}
Training DM using FL enables access to diverse and distributed data, while the challenges of high communication costs and data heterogeneity persist. FedPhD is a first attempt to address these challenges. 
By leveraging Hierarchical FL (HFL), homogeneity-aware aggregation and server selection, 
FedPhD mitigates data heterogeneity while incorporating structured pruning of DMs, thus reducing model size and enhancing computational and communication efficiency. 
Experimental results demonstrate that FedPhD improves the image quality of DMs by at least $34\%$ while achieving approximately $40\%$ reduction in model size, computation, and communication costs. 
Our future agenda includes automatic synchronization of EMA updates in HFL to enhance model convergence balancing computational and communication overhead.

\bibliographystyle{IEEEtran}
\bibliography{ref}

\appendix
\section{Diffusion Models Training}
\begin{algorithm}[H]
\caption{FL-based DM training on client $n$ at round $r$}
\label{alg:client_training_dm}
\begin{algorithmic}[1]
\STATE \textbf{Input:} Global model $\boldsymbol{\theta}^{(r)}$, local dataset $\mathcal{D}_n$, local epochs $E$, batch size $B$, learning rate $\eta$
\STATE \textbf{Output:} Updated local model $\boldsymbol{\theta}^{(r+1)}_n$

\STATE Initialize local model: $\boldsymbol{\theta}_n \leftarrow \boldsymbol{\theta}^{(r)}$
\FOR{epoch $e = 0, \dots, E-1$}
    \FOR{batch $\mathcal{B} \in \mathcal{D}_n$}
        \FOR{each $\mathbf{x}_0 \in \mathcal{B}$ in parallel}
            \STATE Sample $t \sim \text{Uniform}(\{1, 2, \dots, T\})$
            \STATE Compute $\bar{\alpha}_t = \prod_{s=1}^{t} \alpha_s$
            \STATE Sample noise $\boldsymbol{\epsilon} \sim \mathcal{N}(0, \mathbf{I})$
            \STATE Compute noised data $\mathbf{x}_t = \sqrt{\bar{\alpha}_t} \mathbf{x}_0 + \sqrt{1 - \bar{\alpha}_t} \boldsymbol{\epsilon}$
            \STATE Compute $\mathcal{L}(\boldsymbol{\theta}_n)$ using \eqref{eq:noise_predict} 
        \ENDFOR
        \STATE Calculate $F_{n}(\boldsymbol{\theta}_n) = \frac{1}{|\mathcal{B}|} \sum_{\mathbf{x}_0 \in \mathcal{B}} \mathcal{L}(\boldsymbol{\theta}_n)$
        \STATE Update local model: $\boldsymbol{\theta}_n \leftarrow \boldsymbol{\theta}_n - \eta \nabla F_{n}(\boldsymbol{\theta}_n)$
    \ENDFOR
\ENDFOR
\STATE Set $\boldsymbol{\theta}^{(r+1)}_n \leftarrow \boldsymbol{\theta}_n$
\STATE \textbf{return} Updated local model parameters $\boldsymbol{\theta}^{(r+1)}_n$
\end{algorithmic}
\end{algorithm}

\section{Additional Experiments Details}
\subsection{Data Distribution}
\begin{figure}[!ht]
    \centering
    \subfloat[CIFAR-10 Label Distribution\label{fig:cifar10_dist}]{%
        \includegraphics[width=0.35\textwidth]{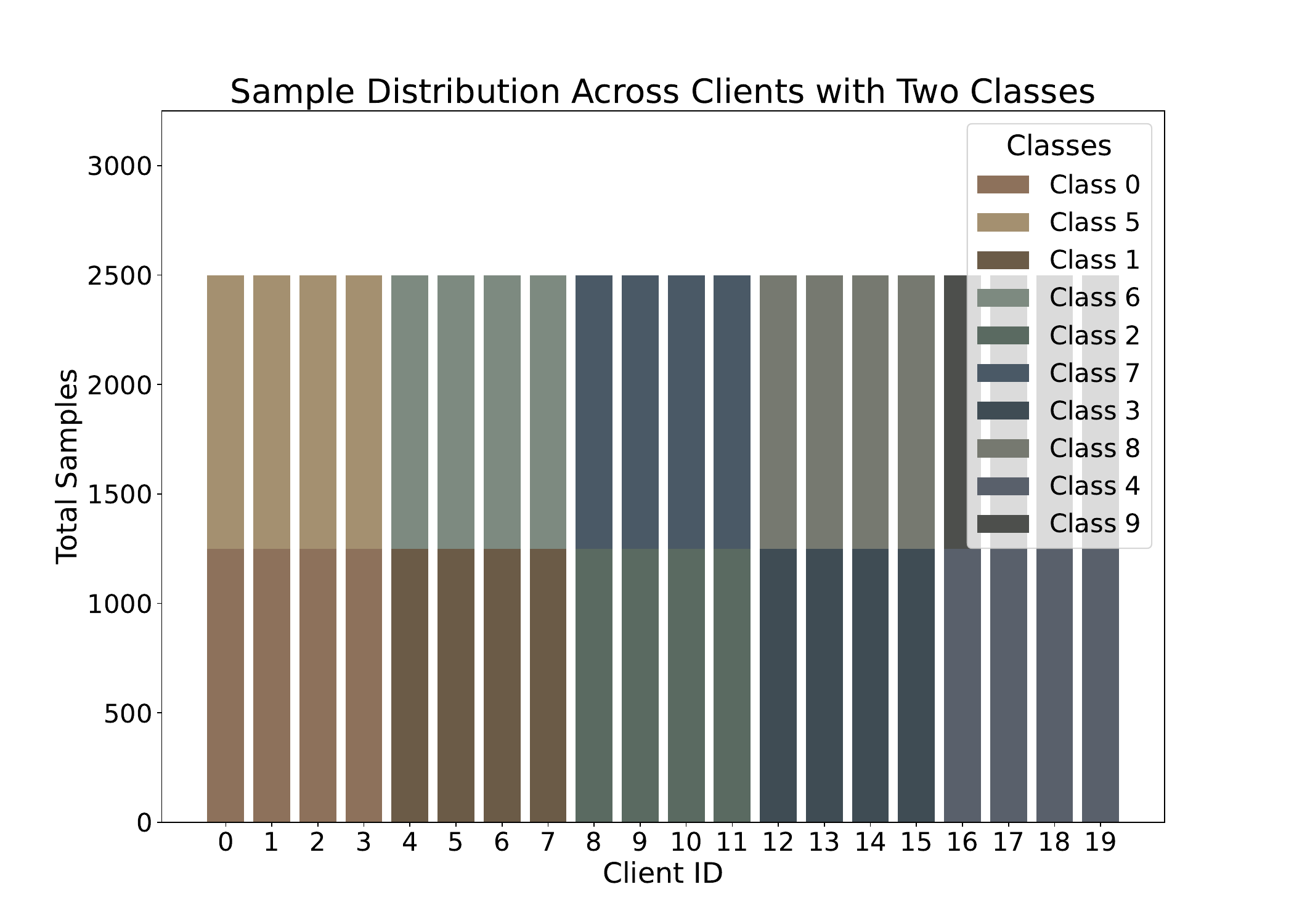}%
    }%
    \hfill
    \subfloat[CelebA Label Distribution\label{fig:celeba_dist}]{%
        \includegraphics[width=0.35\textwidth]{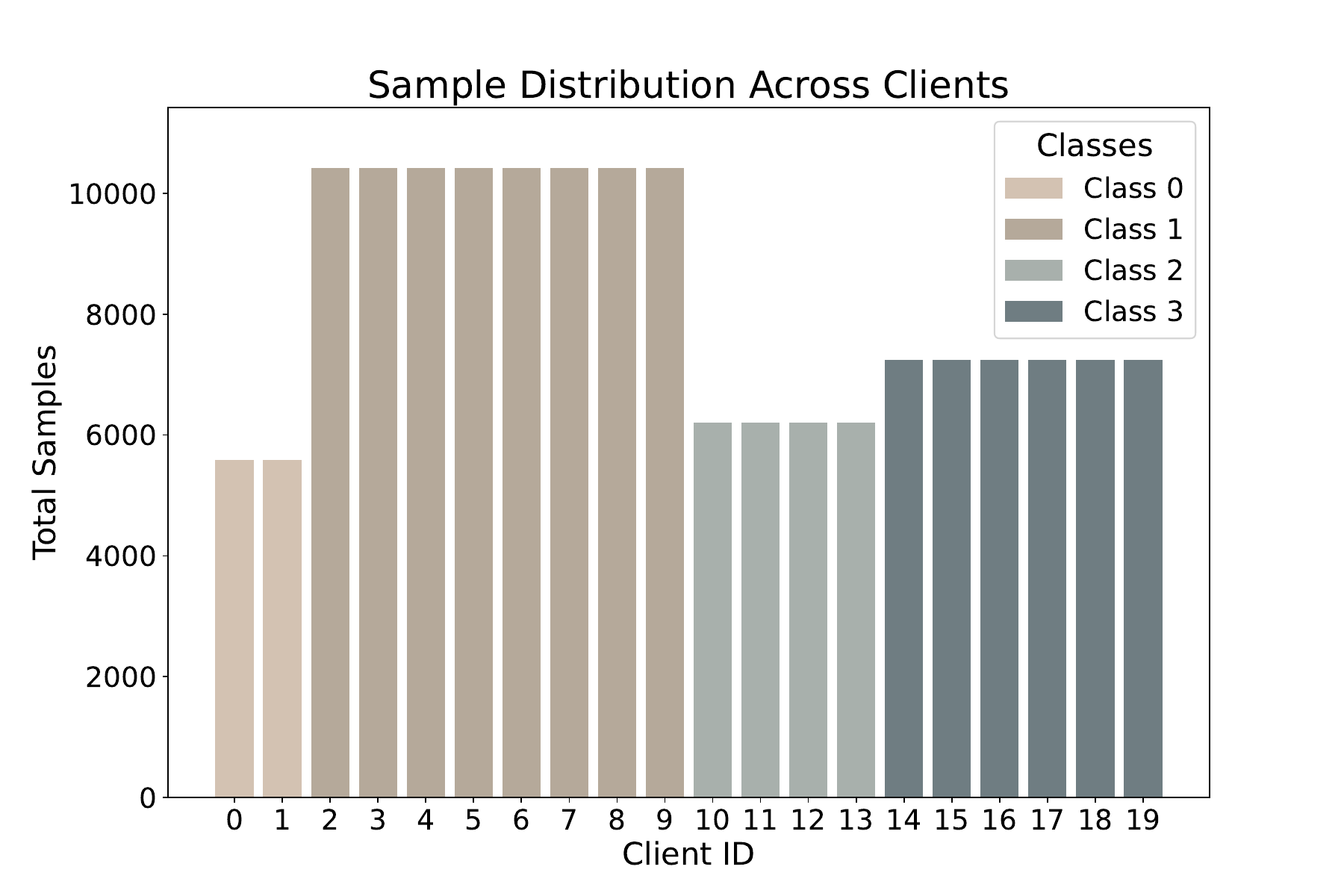}%
    }%
    \caption{Label distribution across clients in CIFAR-10 and CelebA experiments.}
    \label{fig:label_distributions}
\end{figure}

Figure \ref{fig:label_distributions} illustrates the label distributions across clients for the CIFAR-10 and CelebA datasets. The distribution in CIFAR-10 (Figure \ref{fig:cifar10_dist}) exhibits a relatively balanced spread across all classes, whereas the CelebA dataset (Figure \ref{fig:celeba_dist}) demonstrates an inherent imbalance due to its attribute-based classification structure.

\subsection{Generated Images Visualization}
\begin{figure}[!ht]
    \centering
    \subfloat[FID score for CIFAR-10 dataset: 16.74.\label{fig:cifar10_fid}]{%
        \includegraphics[width=0.4\textwidth]{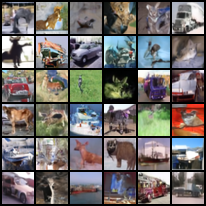}%
    }%
    \hfill
    \subfloat[FID score for CelebA dataset: 7.48.\label{fig:celeba_fid}]{%
        \includegraphics[width=0.4\textwidth]{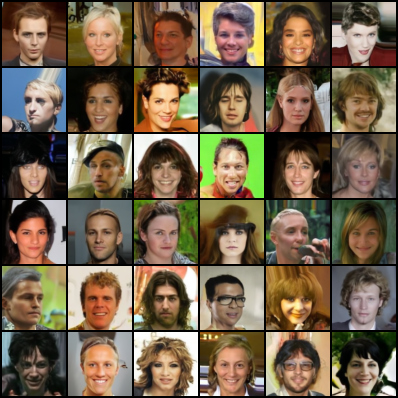}%
    }%
    \caption{Comparison of FID scores for CIFAR-10 and CelebA datasets using FedPhD.}
    \label{fig:fid_comparison}
\end{figure}

Figure \ref{fig:fid_comparison} presents the Fréchet Inception Distance (FID) scores for CIFAR-10 and CelebA datasets when trained using the FedPhD framework. The results show that FedPhD achieves an FID score of 16.74 for CIFAR-10 (Figure \ref{fig:cifar10_fid}) and 7.48 for CelebA (Figure \ref{fig:celeba_fid}). The lower FID score for CelebA suggests a better alignment between real and generated images compared to CIFAR-10, which can be attributed to the structured attribute-based nature of CelebA compared to the object-centric CIFAR-10 dataset. 

\subsection{Baselines}
The baselines adopted are summarized as follows. In distributed training algorithms, we do not apply EMA as it requires frequent synchronization of model updates across nodes, introducing additional computational and communication overhead.
(i) Centralized Training, which serves as a foundational baseline for evaluating the performance difference between distributed and centralized approaches; 
(ii) FedAvg \cite{McMahan2017}, the standard baseline in FL that aggregates model updates from clients; 
(iii) FedProx \cite{Li2020a}, a baseline that balances the trade-off between local computation and communication overhead in FL; 
(iv) FedDiffuse \cite{Goede2024}, a baseline focusing on efficient communication by updating only subsets of DMs; 
(v) MOON \cite{li2021model}, a baseline leverages the similarity between model representations to refine local training for heterogeneity with contrastive learning;
(vi) SCAFFOLD \cite{karimireddy2020scaffold}, a baseline designed to mitigate client drift using control variates that correct model updates in heterogeneous data settings.
\subsection{Hyperparameters Tuning}
The experiments use a batch size of $128$ for CIFAR10 and $64$ for CelebA, with learning rates of $2 \times 10^{-4}$ and $2 \times 10^{-5}$, respectively. The optimizer employed is Adam. The balance parameter $a$ is determined through a grid search in the range $[0, 50K]$, while $b$ is chosen to ensure that the numerator in (\ref{eq:edge_server_selection}) remains positive. The regularization factor $\lambda_{g}$ for group norm pruning is tuned via grid search, ranging from $10^{-2}$ to $10^{-5}$, depending on the task. For FedProx, we set the regularization parameter to 1, following the recommendation in \cite{Li2020a}. The weight for the contrastive loss in MOON is determined through a grid search of the values 
[0.01,0.1,1,10]. For FedPhD, the sparse training rounds $R_{s} \in [10,100]$ is set by using grid search.

\begin{table}[!ht]
\centering
\caption{Scalability Evaluation on CIFAR10.}
\begin{tabular}{lcccccc}
\hline
\multirow{2}{*}{Method} & \multicolumn{2}{c}{$N=50$} & \multicolumn{2}{c}{$N=100$} \\
\cline{2-3} \cline{4-5}
& FID $\downarrow$ & IS $\uparrow$ & FID $\downarrow$ & IS $\uparrow$ \\
\hline
FedAvg & 25.48 & 4.01 & 23.11 & 4.08\\
FedProx & 26.22 & 3.98 & 22.88 & 4.04 \\
FedDiffuse & 28.84 & 3.88 & 25.22 & 4.01 \\
SCAFFOLD & 49.23 & 3.09 & 45.44 & 3.11 \\
MOON     & 22.67    & 4.04     & 23.24     & 4.08  \\
FedPhD  & \textbf{17.14} & \textbf{4.36} & \textbf{17.77} & \textbf{4.11} \\
FedPhD (OS) & 17.62 & 4.18 & 18.03 & 4.09\\
\hline
\end{tabular}
\label{tab:scalability_metrics}
\end{table}
\subsection{Scalability Performance}
In the main experiments, we consider the setting of $N=20$ clients. 
We further examined the impact of the number of clients $N$ on the baselines using CIFAR10 with $N=50$ and $N=100$. 
The results in Table \ref{tab:scalability_metrics} illustrate that the performance of DDIM trained in a distributed manner deteriorates as the number of clients increases from $N=20$ to $N=50$, despite the total number of images remaining constant. 
In this case, the impact on \textit{FedPhD} is lower compared to other baselines. Less frequent aggregation allows for more local computation, which causes models to become `biased', especially as the number of clients $N$ increases and the data per client decreases. 
This amplifies the data heterogeneity issue, 
leading to over-fitting on smaller, non-representative datasets and greater divergence during aggregation. Alternatively, more frequent aggregation mitigates this by regularly synchronizing model updates, preventing local over-fitting, and keeping the global model closer to the overall data distribution. 
In terms of the quality of the generated images, there is only a $0.4$ increase in FID for \textit{FedPhD}, while the increase is around $3.5$ for other baselines. Similar results are obtained with $N=100$.

\section{Edge Server Selection in FedPhD}
\begin{figure}[ht!]
    \centering
    \includegraphics[width=0.5\textwidth]{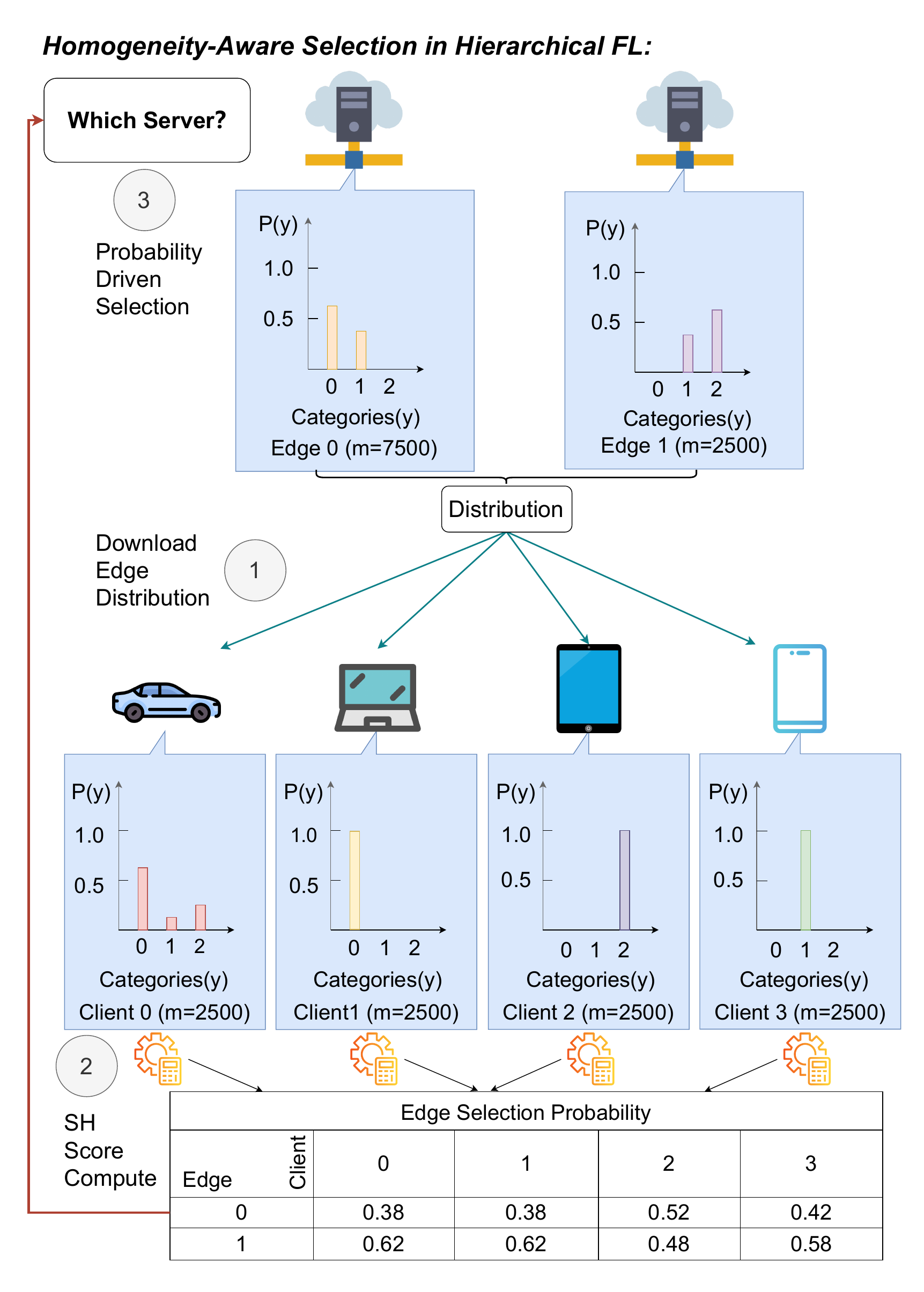}
    \caption{Example of Homogeneity-Aware Edge Server Selection in HFL. 
    The edge selection probabilities, calculated using \eqref{eq:edge_server_selection} with parameters $a = 15000$ and $b = 0$, are provided for each client based on the SH. Compared to \textit{random selection}, an edge server with a higher SH or fewer attached samples has a higher probability of being selected.}
    \label{fig:edge_server_selection}
\end{figure}

First, we provide an example of the procedure of edge server selection in Figure \ref{fig:edge_server_selection}.

\begin{figure*}[!ht] 
    \centering
    \includegraphics[width=0.8\textwidth]{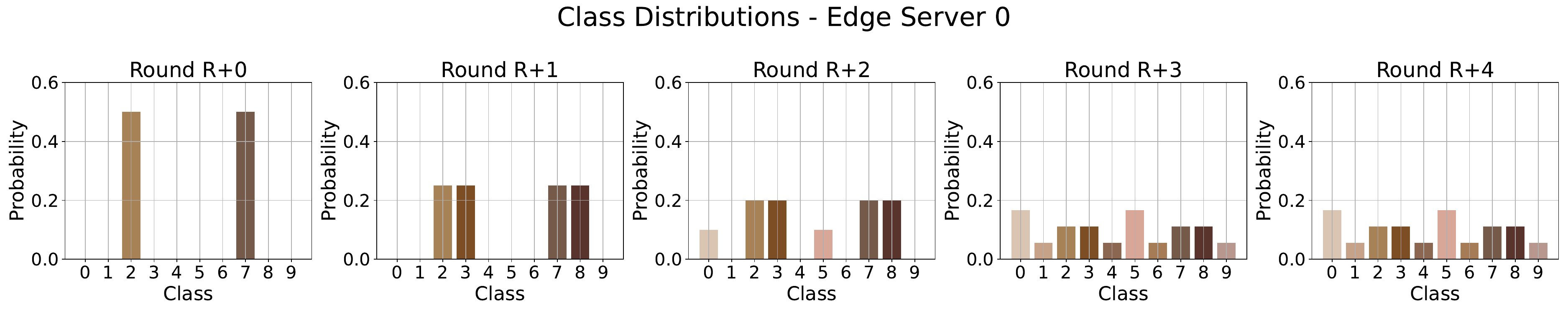}
    \hfill
    \includegraphics[width=0.8\textwidth]{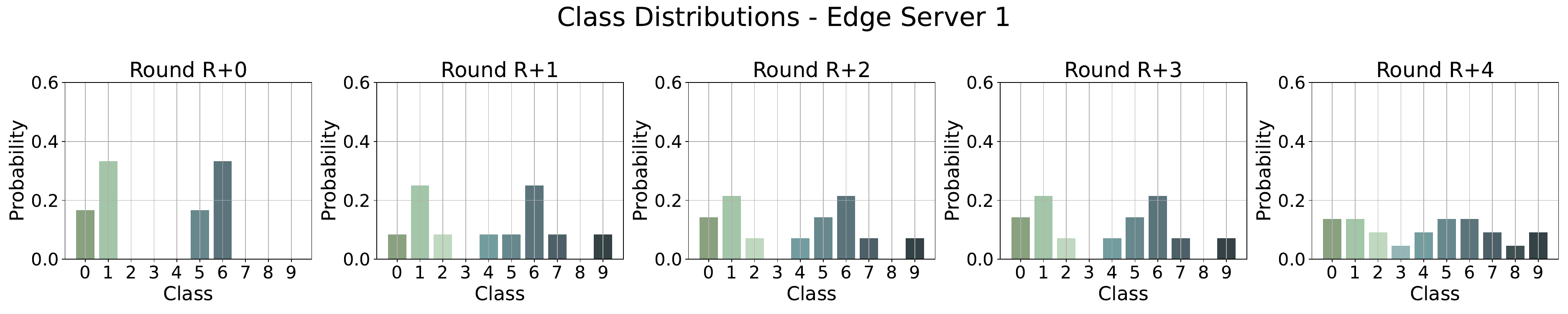}
    \caption{Accumulated class distributions for Edge Server 0 and Edge Server 1 (CIFAR10); Homogeneity-Aware Selection.}
    \label{fig:homo_selection}
\end{figure*}

\begin{figure*}[!ht] 
    \centering
    \includegraphics[width=0.8\textwidth]{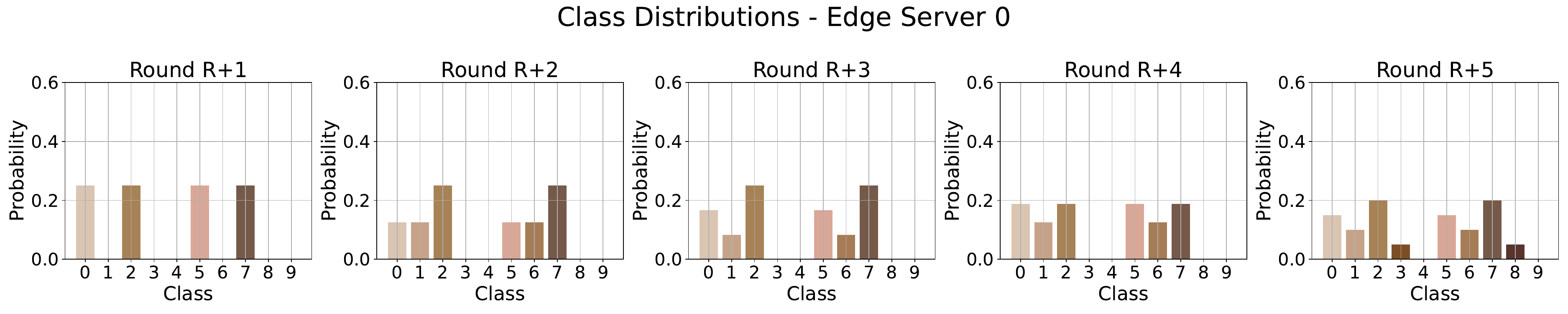}
    \hfill
    \includegraphics[width=0.8\textwidth]{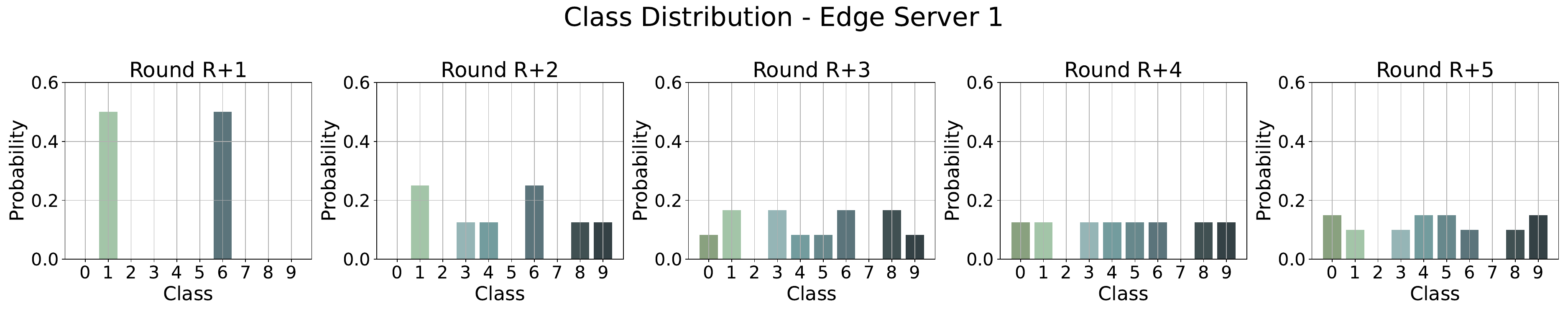}
    \caption{Accumulated class distributions for Edge Server 0 and Edge Server 1 (CIFAR10); Random Selection.}
    \label{fig:random_selection}
\end{figure*}

Figures \ref{fig:homo_selection} and \ref{fig:random_selection} show the proposed edge server selection mechanism and the random selection method, respectively (similar results are obtained with more edge servers; not shown here due to space limitations). 
From round $R$, divisible by $r_{g}=5$, to round $R+4$, the accumulated distribution changes over the edge servers are evident. 
To fairly demonstrate the advantages of FedPhD's selection mechanism, we compute the final SH score $\mu_{e}$ at round $R+4$ for both FedPhD and random selection. The SH scores for FedPhD are $1.869$ and $1.896$ for edge servers 0 and 1, respectively, while the random selection method yields scores of $1.816$ and $1.811$. 
Our method outperforms random selection, producing more homogeneous class distributions across edge servers, indicating improved balance while reducing data skewness.

\begin{figure}[!ht] 
    \centering
    \includegraphics[width=0.45\textwidth]{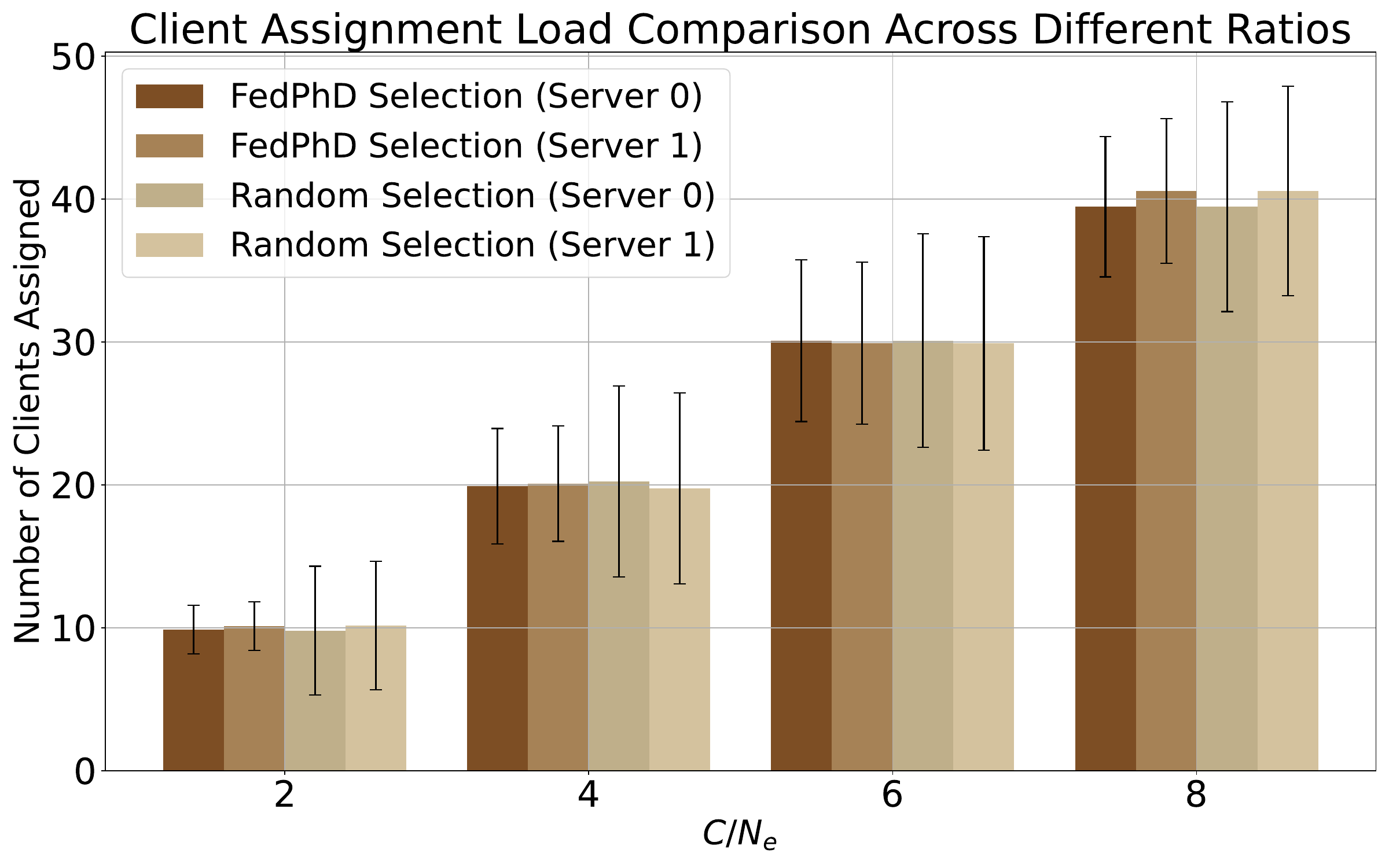}
    \caption{Load comparison in terms of number of clients assigned to edge servers (every $r_{g} = 5$ rounds).}
    \label{fig:load_comparison}
\end{figure}

The client assignment analysis in Fig. \ref{fig:load_comparison} highlights the effectiveness of FedPhD in maintaining balanced client distributions compared to random selection. 
The variance depends on the ratio $\frac{C}{N_{e}}$ representing the average number of assigned clients per edge server. We present results for $N=20$ clients with ratios $\{2, 4, 6, 8\}$. 
For instance, with $\frac{C}{N_{e}}=2$, FedPhD achieves an average of $9.88$ and $10.12$ assigned clients to edge servers 0 and 1, respectively, resulting in variance $0.72$. 
Random selection exhibits significantly higher variance in client assignments. The reduced variance in FedPhD not only indicates consistently balanced client assignments but also minimizes disparities and ensures uniform workload. This stability contributes to improved model convergence rates and efficient edge server resource utilization, as opposed to uneven resource usage observed with random selection in traditional FL. Finally, FedPhD identifies potential candidates for edge server selection in the event that an edge server fails to operate. By ranking edge servers $e$ based on probability contribution $P_{n}(e)$ to SH score, the $k^{th}$ best server acts as the alternative one for selection by client $n$ ensuring resilience, maximizing SH and maintaining system performance.

\end{document}